\documentclass[10pt, twocolumn, letterpaper]{article}

\usepackage[pagenumbers]{cvpr} % To force page numbers, e.g. for an arXiv version

\definecolor{cvprblue}{rgb}{0.21,0.49,0.74}
\usepackage[pagebackref, breaklinks, colorlinks, allcolors=cvprblue]{hyperref}
\usepackage{multirow}
\usepackage{makecell}

\def\paperID{2128} % *** Enter the Paper ID here
\def\confName{CVPR}
\def\confYear{2025}

\title{\textit{SpikePack}: Enhanced Information Flow in Spiking Neural Networks\\ with
High Hardware Compatibility}

\author{
	Guobin Shen$^{1, 2}$, \ Jindong Li$^{1, 3}$, 
    \ Tenglong Li$^{1, 3}$ \ Dongcheng Zhao$^{1}$, \ and Yi Zeng$^{1, 2, 3,}$\footnotemark[2] \\
    \normalsize$^1$ BrainCog Lab, CASIA, 
    $^2$ School of Future Technology, UCAS, 
    $^3$ School of Artificial Intelligence, UCAS \\
    \small\texttt{\{shenguobin2021, lijindong2022, litenglong2023, zhaodongcheng2016, yi.zeng\}@ia.ac.cn}
}

\begin{document}
    \maketitle
    \begin{abstract}
    Spiking Neural Networks (SNNs) hold promise for energy-efficient,
    biologically inspired computing. We identify substantial information loss during
    spike transmission, linked to temporal dependencies in traditional Leaky
    Integrate-and-Fire (LIF) neurons—a key factor potentially limiting SNN
    performance. Existing SNN architectures also underutilize modern GPUs,
    constrained by single-bit spike storage and isolated weight-spike operations
    that restrict computational efficiency. We introduce \textit{SpikePack}, a neuron
    model designed to reduce transmission loss while preserving essential features
    like membrane potential reset and leaky integration. \textit{SpikePack} achieves
    constant $\mathcal{O}(1)$ time and space complexity, enabling efficient parallel
    processing on GPUs and also supporting serial inference on existing SNN
    hardware accelerators. Compatible with standard Artificial Neural Network (ANN)
    architectures, \textit{SpikePack} facilitates near-lossless ANN-to-SNN conversion
    across various networks. Experimental results on tasks such as image
    classification, detection, and segmentation show \textit{SpikePack} achieves
    significant gains in accuracy and efficiency for both directly trained and converted
    SNNs over state-of-the-art models. Tests on FPGA-based platforms further
    confirm cross-platform flexibility, delivering high performance and enhanced
    sparsity. By enhancing information flow and rethinking SNN-ANN integration, \textit{SpikePack}
    advances efficient SNN deployment across diverse hardware platforms.
\end{abstract}
    \vspace{-6mm}
\section{Introduction}
\label{sec:intro}

Spiking Neural Networks (SNNs)~\cite{maass1997networks} have emerged as a promising
paradigm for energy-efficient and biologically inspired computing~\cite{zhang2020system}.
By emulating the discrete spike-based communication of biological neurons, SNNs offer
potential advantages in terms of low-power consumption and event-driven processing,
which are particularly appealing for deployment on neuromorphic hardware~\cite{merolla2014million,pande2010embrace,
li2023firefly, li2024firefly, li2024fireflys}.

Despite these advantages, SNNs still face significant challenges that impede their
widespread application in complex tasks such as image classification~\cite{wu2018spatio,zhou2022spikformer,liu2018rethinking,
shen2022backpropagation, zhao2025improving}, object detection~\cite{luo2024integer,
yaospiked, li2022spike}, and natural language processing~\cite{shen2023astrocyte,
xing2024spikellm}. Notably, their performance often lags behind that of
Artificial Neural Networks (ANNs). One key reason we have identified is the
substantial information loss that occurs during spike transmission, particularly
associated with the temporal dependencies inherent in traditional neuron models like
the Leaky Integrate-and-Fire (LIF) neuron~\cite{dayan2005theoretical}. This
information degradation can limit the network's ability to capture and transmit
critical features, thereby hindering overall performance.

\begin{figure}[t]
  \centering
  \includegraphics[width=\linewidth]{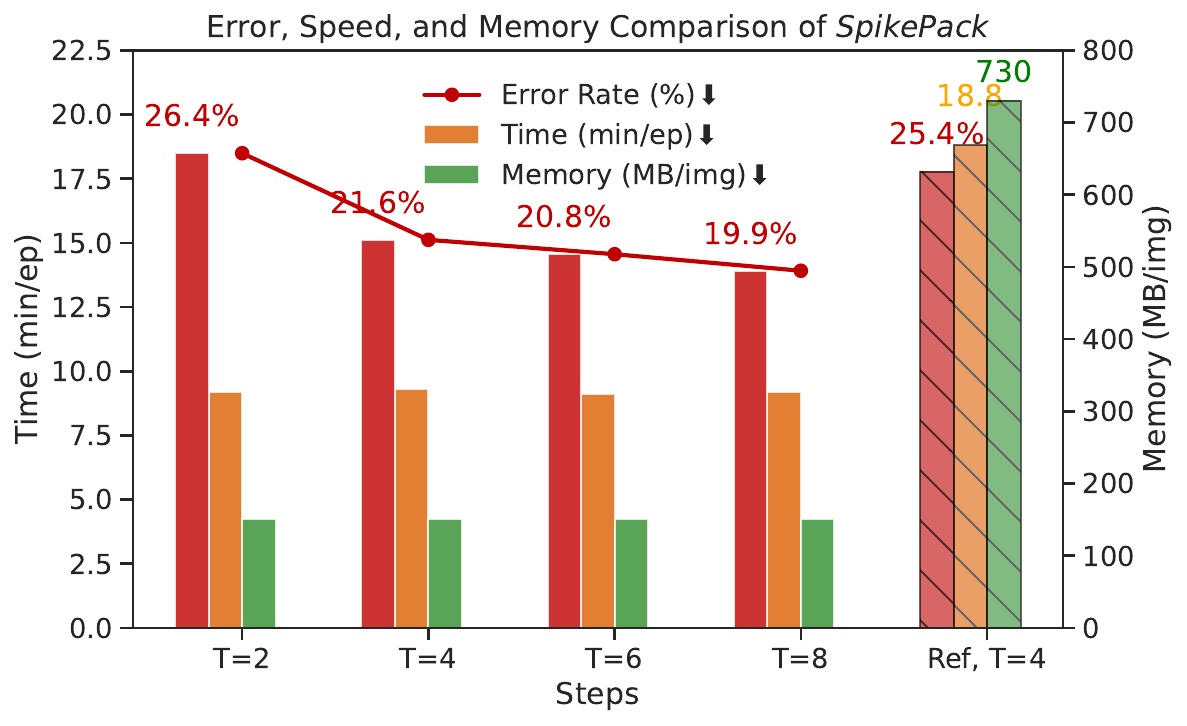}
  \caption{Performance of \textit{SpikePack} on Spikeformer-8-512 across different time steps (T), showing error rate (\%), time (ep/min), and memory (MB/img). Reference model uses LIF neurons (Ref, T=4).}
  \label{fig:overview}
  \vspace{-6mm}
\end{figure}

Moreover, existing SNN architectures have not fully exploited the capabilities
of modern General-Purpose Graphics Processing Units (GPGPUs). The reliance on single-bit
spike representations~\cite{shen2023exploiting} and isolated weight-spike operations~\cite{shen2024conventional,luo2024integer}
leads to inefficient utilization of parallelism, resulting in low training
efficiency and slow inference speeds. This inefficiency not only hampers the practicality
of SNNs but also complicates their deployment across diverse hardware platforms~\cite{carpegna2024spiker,
kuwahara2024fusion}.

Although there has been considerable research into neuromorphic hardware~\cite{pei2019towards}
and the development of various SNN accelerators~\cite{li2024firefly, fang2023spikingjelly},
SNNs have not become mainstream. This is partly due to their subpar performance compared
to ANNs and their incompatibility with modern ANN architectures~\cite{fang2021deep,
zhou2022spikformer}. This incompatibility arises not only from the reliance on
discrete spikes but also from the temporal dependencies in spiking neurons, which
require SNN-specific network designs and training methods. Consequently, adapting
ANN models and techniques to SNNs requires complex modifications, limiting SNNs'
ability to fully leverage advancements in ANN architectures and optimization.

To address these challenges, we propose \textit{SpikePack}, a novel neuron model
designed to reduce information loss during the transition from pre-synaptic to
post-synaptic spikes, minimizing degradation associated with temporal dependencies
in traditional neuron models. \textit{SpikePack} enhances information flow within
SNNs while achieving $\mathcal{O}(1)$ time and space complexity with respect to time
steps, enabling efficient time-parallel training and inference on GPUs, as shown in Figure~\ref{fig:overview}. \textit{SpikePack} also preserves essential biological characteristics, such as membrane potential
reset and leaky integration, ensuring biological plausibility.

In addition, \textit{SpikePack} is compatible with modern ANN architectures, allowing
for near-lossless ANN-to-SNN conversion and preserving the inherent sparsity of spike-based
computations. This compatibility enables the integration of advanced ANN models
within the SNN framework, improving performance across a variety of tasks.

\begin{figure}[h]
    \centering
    \includegraphics[width=\linewidth]{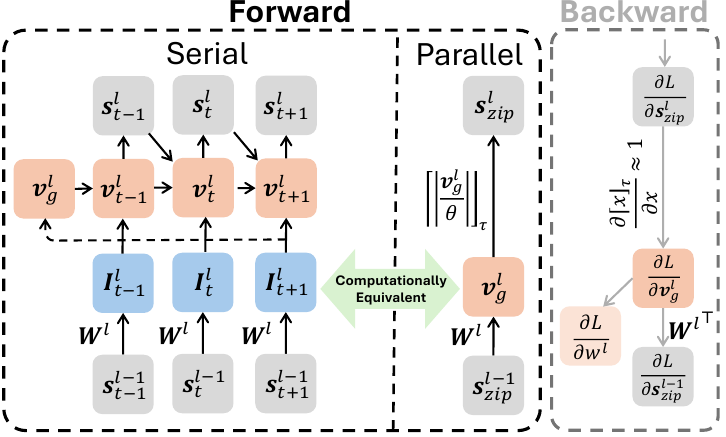}
    \caption{Forward and backward computation in \textit{SpikePack}, showing both serial and parallel modes. Parallel computation uses \( s_{\text{zip}}^l \) for efficient global potential (\( v_g^l \)) calculation.}
    \label{fig:forward_backward_computation}
    \vspace{-4mm}
\end{figure}

Our contributions can be summarized as follows:

\begin{itemize}
  \item We introduce \textit{SpikePack}, a neuron model that minimizes information
    loss from pre-synaptic to post-synaptic spikes. \textit{SpikePack} achieves a
    balance between computational efficiency and biological fidelity in SNNs.

  \item By achieving $\mathcal{O}(1)$ time and space complexity, \textit{SpikePack} preserves
    essential neural dynamics, enabling efficient, biologically relevant behavior. As shown in Figure~\ref{fig:forward_backward_computation}, \textit{SpikePack} supports
    direct training, eliminating the need for complex
    temporal unfolding and enabling more efficient gradient-based optimization. Its
    compatibility with modern ANN architectures further supports near-lossless ANN-to-SNN
    conversion while maintaining the inherent sparsity of spike-based computations.

  \item Extensive experiments on image classification, object detection, and segmentation
    tasks showcase significant improvements over state-of-the-art methods.
    Additional testing on SNN hardware accelerators further validates the
    generality and efficiency of our approach.
\end{itemize}

By addressing the fundamental issues of information loss and hardware
inefficiency, \textit{SpikePack} represents a significant advancement in the
practical deployment of SNNs. It not only enhances the computational capabilities
of SNNs but also ensures that these improvements are accessible across various hardware
platforms. This work not only brings us closer to realizing the full potential
of neuromorphic computing in real-world scenarios but also offers new insights
into bridging SNNs and ANNs.

\section{Related Work}
\label{2_related_work}

Our work builds on advancements in three areas of SNNs: information transmission efficiency, efficient training methods, and compatibility with ANNs.

\textbf{Information Transmission Efficiency} Traditional neuron models like the LIF neuron rely on binary spikes, which limit information capacity. To address this, ternary spike neurons~\cite{guo2024ternary} and burst-based LIFB neurons~\cite{shen2023exploiting} expand spike representations, allowing richer information flow while retaining energy efficiency. Other approaches, such as integer-valued neurons~\cite{luo2024integer} and rectified membrane potentials~\cite{guo2022recdis}, aim to reduce quantization errors and mitigate degradation in deep SNNs. Although these methods improve information transmission, they often add model complexity. Our \textit{SpikePack} offers a simpler approach by computing membrane potentials before spikes are generated, effectively enhancing information flow without increasing computational demands.

\textbf{Efficient Training Methods} Training efficiency is a core challenge in SNNs due to the non-differentiability of spike operations, and weight-spike computations, all of which limit hardware utilization. While Backpropagation Through Time (BPTT)~\cite{wu2018spatio} has enabled deep SNN training, it incurs high memory and time costs. Approaches like Online Training Through Time (OTTT)~\cite{xiao2022online} and Spatial Learning Through Time (SLTT)~\cite{meng2023towards} reduce memory overhead by prioritizing critical temporal interactions, while Temporal Reversible SNNs (T-RevSNN)~\cite{huhigh2024} leverage reversible architectures to lower memory costs. However, these methods primarily optimize training efficiency without addressing the fundamental limitations of binary spikes and underutilized hardware. In contrast, \textit{SpikePack} inherently supports time-parallel processing, achieving \( \mathcal{O}(1) \) complexity with respect to time steps, thus improving training efficiency and enabling high inference efficiency across diverse hardware.

\textbf{Compatibility with ANN Architectures} Leveraging ANN advancements within the SNN framework is challenging due to fundamental differences in activation dynamics. Traditional ANN-to-SNN conversion approaches rely on rate coding, often requiring many time steps and recalibration~\cite{li2022spike}. Recently, methods such as Spatio-Temporal Approximation~\cite{jiang2024spatio} and Expectation Compensation~\cite{huang2024towards} have enabled SNN adaptations of Transformer architectures by approximating non-linear interactions. While effective, these methods are often complex and architecture-specific. By contrast, \textit{SpikePack} enables near-lossless conversion across various ANN architectures with minimal adjustments, facilitating direct integration with modern ANN models while preserving SNN sparsity and efficiency.

While previous works have addressed aspects of information flow, training efficiency, and ANN compatibility, they often require complex modifications. \textit{SpikePack} provides a unified, streamlined solution that enhances information flow, supports efficient training, and enables seamless integration with ANN architectures, promoting scalable SNN deployment across diverse hardware.

    \section{Methodology}
\label{sec:method}

In this section, we analyze the limitations of LIF neurons, focusing on their information transmission inefficiencies and computational limitations on GPGPUs. These limitations stem from both information loss during spike transmission and inefficient hardware utilization. To address these challenges, we propose \textit{SpikePack}, a novel neuron model designed to preserve critical information and support efficient parallel processing on modern hardware. We also provide a theoretical foundation for \textit{SpikePack} through analysis of information transmission, followed by an explanation of its computational efficiency.

\begin{figure}[hb]
    \centering
    \includegraphics[width=.9\linewidth]{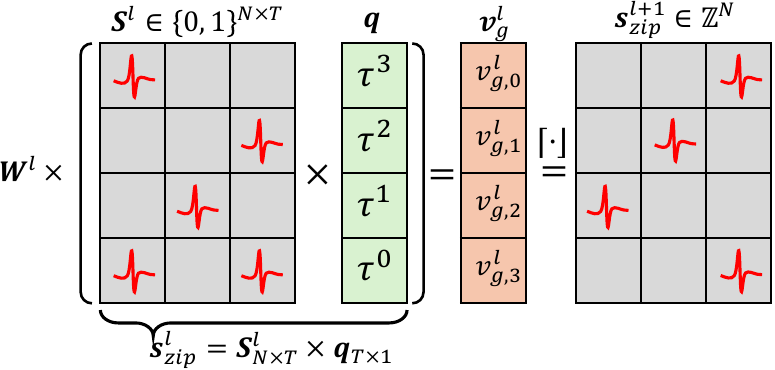}
    \caption{Spike sequence compression into integer \( s_{\text{zip}}^l \) for efficient computation of \( v_g^l \) without decompression.}
    \label{fig:spike_sequence_compression}
    \vspace{-6mm}
\end{figure}

\subsection{Limitations of LIF Neurons}
\label{sec:lif_limitations}

The LIF neuron model is widely used in SNNs due to its simplicity and biological inspiration. However, it suffers from two key limitations that hinder its effectiveness: low information retention during spike transmission and inefficient utilization of modern parallel processing hardware.

\subsubsection{Low Information Capacity in Spike Transmission}
\label{sec:lif_info_capacity}

LIF neurons generate output spikes based on the membrane potential at each discrete time step. However, since the membrane potential is determined by the combined effects of the historical input sequence and the decay factor, the model exhibits certain limitations in fully integrating input information. This means that each spiking decision is based on an incomplete view of the input sequence, causing substantial information loss.

For a given layer \( l \) in an SNN, let \( \mathbf{S}^{l-1} \in \{0, 1\}^{N \times T} \) denote the binary spike matrix from the previous layer, where \( N \) is the number of pre-synaptic neurons, and \( T \) is the number of time steps. The weight matrix of layer \( l \) is represented as \( \mathbf{W}^{l} \in \mathbb{R}^{M \times N} \), where \( M \) is the number of neurons in the current layer. For our analysis, we focus on a single neuron in this layer, represented by the weight vector \( \mathbf{w}^{l} = \mathbf{W}_{i, :}^{l} \). Here, we refer to the output spike sequence for this neuron as \( \mathbf{s}^{l} \in \{0, 1\}^{1 \times T} \).

The membrane potential \( v_{t} \) of a LIF neuron updates according to the rule given in Eq.~\eqref{eq:lif_membrane}:
\begin{equation}
    v_{t}^l= \frac{1}{\tau}v^l_{t-1}+ \mathbf{w}^{l} \cdot \mathbf{s}_{t}^{l-1} - \theta \cdot s_{t-1}^{l}, \label{eq:lif_membrane}
\end{equation}
where \( \tau \) is the membrane time constant, \( \mathbf{s}_{t}^{l-1} \) represents input spikes at time \( t \), and \( \theta \) denotes the firing threshold. A spike is generated when \( v_{t} \) exceeds the threshold \( \theta \), after which the membrane potential is reset to facilitate subsequent spiking dynamics as shown in Eq.~\eqref{eq:lif_spike}:
\begin{equation}
    s_{t}^{l}=
    \begin{cases}
        1, & \text{if } v_{t} > \theta, \\
        0, & \text{otherwise}.
    \end{cases}
    \label{eq:lif_spike}
\end{equation}

Since \( v_{t}^l \) is recursively dependent on \( v_{t-1}^l \), each spiking decision is based on partially integrated information from the input sequence. This recursive structure imposes inherent limitations on the mutual information that can be preserved between the input and output, leading to significant information loss during spike transmission. As a result, the ability of the model to fully encode and utilize temporal dependencies in the input signal is constrained.

\subsubsection{Inefficient Hardware Utilization}
\label{sec:lif_hardware}

Another significant limitation of LIF neurons lies in their inefficient utilization of modern parallel processing hardware. The recursive nature of the membrane potential update, as defined in Eq.~\eqref{eq:lif_membrane}, inherently restricts parallelism, as each time step must be processed sequentially, resulting in a time complexity of \( \mathcal{O}(T) \), where \( T \) is the number of time steps. Furthermore, the need to store individual spikes for each time step occupies separate integer or floating-point units in memory, leading to inefficient memory utilization and increased space complexity of \( \mathcal{O}(T) \). These inefficiencies are especially problematic for hardware architectures optimized for parallel computation.

In addition, the LIF model requires repeated spike-weight operations at every time step. As each input at each time step consists of a single spike value, the effective computation is limited to spike-weight multiplications, thereby underutilizing the General Matrix Multiplication (GeMM) capabilities of modern hardware. During training, the reliance on surrogate gradient approximations~\cite{wu2018spatio} to handle the non-differentiable spiking functions further exacerbates computational inefficiency, increasing both the computational overhead and overall complexity.

\subsection{\textit{SpikePack}}
\label{sec:spikepack_neuron}

To address these challenges, we introduce \textit{SpikePack}, a neuron model designed to preserve information capacity across the input sequence while supporting efficient parallel computation. \textit{SpikePack} aggregates information across the entire input sequence into a global membrane potential, denoted as \( v_{g}^{l} \) for layer \( l \). This global aggregation enhances information flow and maximizes mutual information between inputs and outputs.

The global membrane potential \( v_{g}^{l} \) is computed as:
\begin{equation}
    v_{g}^{l} = \mathbf{w}^{l} \mathbf{S}^{l-1} \mathbf{q}, \label{eq:spikepack_global_potential}
\end{equation}
where \( \mathbf{q} = [\tau^{T-1}, \tau^{T-2}, \dots, \tau^{0}]^{\top} \) applies the influence of the membrane time constant \( \tau \) across time steps. This formulation enables \( v_{g}^{l} \) to integrate information from the entire input sequence \( \mathbf{S}^{l-1} \), resulting in improved information flow from pre-synaptic to post-synaptic neurons.

After aggregating information into \( v_{g}^{l} \), \textit{SpikePack} generates the output spike sequence \( \mathbf{S}^{l} \) through a decoding process. The membrane potential \( v_{t}^l \) is updated as in Eq.~\eqref{eq:spikepack_reset}:
\begin{equation}
    v_{t}^l = v_{t-1}^l - \theta_{t} \cdot s_{t-1}^{l}, \label{eq:spikepack_reset}
\end{equation}
where \( \theta_{t} = \frac{\theta}{\tau^{t - T}} \) represents a dynamic threshold that adapts over time. The initial membrane potential is set to \( v_0^l = v_g^l \). The spike generation condition is defined as in Eq.~\eqref{eq:spikepack_spike}:
\begin{equation}
    s_{t}^{l}=
    \begin{cases}
        1, & \text{if } v_{t}^l > \theta_{t}, \\
        0, & \text{otherwise}.
    \end{cases}
    \label{eq:spikepack_spike}
\end{equation}

% By deferring the generation of spikes until after the aggregation into \( v_{g}^{l} \), \textit{SpikePack} achieves higher mutual information, allowing more precise transmission of information to subsequent layers.

\subsubsection{Improved Information Capacity in \textit{SpikePack}}

\begin{figure}[b]
    \centering
    \includegraphics[width=.85\linewidth]{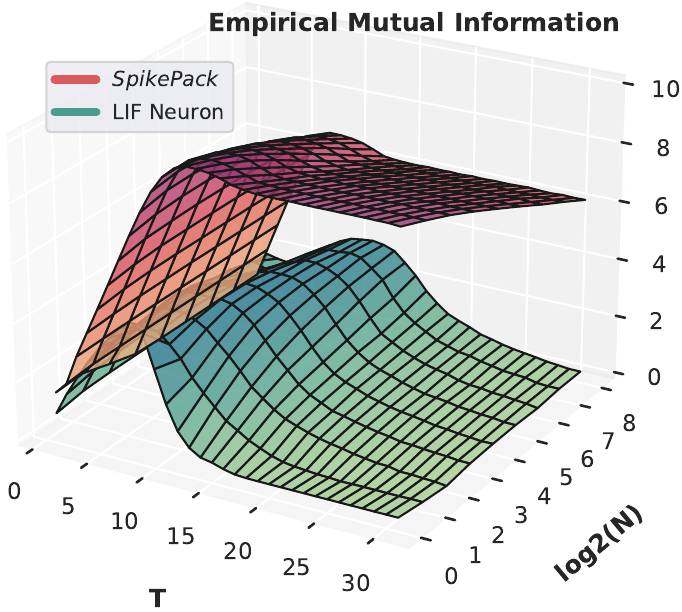}
    \vspace{-2mm}
    \caption{Empirical comparison of mutual information between \textit{SpikePack} and
    LIF neurons over varying $T$ (time steps) and $N$ (pre-synaptic neurons).
    \textit{SpikePack} demonstrates higher information retention across configurations.}
    \label{fig:mi}
    \vspace{-2mm}
\end{figure}

The initial global membrane potential \( v_{g}^{l} \) aggregates information across the entire input sequence, thereby enhancing the mutual information between the input \( \mathbf{S}^{l-1} \) and the output spike sequence \( \mathbf{s}^{l} \). Assuming binary spikes with independent Bernoulli distributions and Gaussian weights, \( v_{g}^{l} \) approximates a Gaussian distribution with variance as shown in Eq.~\eqref{eq:variance_spikepack}:

\begin{equation}
    \sigma_{v_{g}^{l}}^{2} = \sigma^{2} N p (1 - p) \left( \sum_{t=1}^{T} q_{t} \right)^{2},
    \label{eq:variance_spikepack}
\end{equation}
where \( p \) represents the probability of an input spike, and \( q_{t} = \tau^{t-1} \).

The mutual information between \( \mathbf{S}^{l-1} \) and \( \mathbf{s}^{l} \) for \textit{SpikePack} can thus be approximated as in Eq.~\eqref{eq:mi_spikepack_final}:
\begin{equation}
    I_{\mathrm{SP}}^{l} = \frac{1}{2} \log_{2} \left( \frac{12 \sigma_{v_{g}^{l}}^{2}}{\theta^{2}} \right).
    \label{eq:mi_spikepack_final}
\end{equation}
This result, derived theoretically and validated empirically, demonstrates that \textit{SpikePack} achieves greater information retention across a range of configurations for \( N \) and \( T \), affirming its superior transmission capacity.

For a more detailed theoretical derivation of the mutual information \( I(\mathbf{S}_{\mathrm{in}}, \mathbf{s}_{\mathrm{out}}) \) between pre-synaptic and post-synaptic spikes in both the LIF and \textit{SpikePack} models, please refer to Appendix~\ref{appendix:mutual_information}. Empirical simulations, as illustrated in Figure~\ref{fig:mi}, confirm that \textit{SpikePack} consistently achieves higher mutual information across various values of \( N \) and \( T \), further substantiating its enhanced information transmission capability.

\subsubsection{Parallel Computation and Hardware Utilization}
\label{sec:spikepack_parallel}

The \textit{SpikePack} model leverages the compressed input structure \( v_{g}^{l} \), making its operations hardware-friendly and well-suited for parallel computation. As shown in Figure~\ref{fig:spike_sequence_compression}, since all spike information is aggregated into \( v_{g}^{l} \), both time and space complexity are reduced from \( \mathcal{O}(T) \) to \( \mathcal{O}(1) \), enabling efficient use of parallel processing on modern GPUs.

The packed representation of input spikes enables highly efficient matrix multiplication operations, thereby optimizing computational performance. The global membrane potential \( v_{g}^l \) is computed as in Eq.~\eqref{eq:v0_compressed}:
\begin{equation}
    v_{g}^l = \mathbf{w}^{l} \mathbf{s}^{l-1}_\mathrm{zip}, \label{eq:v0_compressed}
\end{equation}
where \( \mathbf{s}^{l-1}_\mathrm{zip} = \mathbf{S}^{l-1} \mathbf{q} \). This formulation compresses the presynaptic spike matrix into a bitwise representation, facilitating optimized GeMM operations and enhancing computational efficiency.

Utilizing this compressed representation, the total output spike count over \( T \) time steps can be derived directly from \( v_{g}^l \) without iterative updates, as in Eq.~\eqref{eq:so_zipped}:
\begin{equation}
    s_{\mathrm{zip}}^l = \left\lceil \frac{v_{g}^l}{\theta} \right\rfloor_{\tau} . \label{eq:so_zipped}
\end{equation}

In this representation, each bit in \( s_{\mathrm{zip}}^l \) indicates whether an output spike \( s_{t}^l \) is generated at time step \( t \), effectively compressing the spike sequence. This compact form aligns with the serial spike generation in Eqs.~\eqref{eq:spikepack_reset} and \eqref{eq:spikepack_spike}, preserving spiking behavior without per-step computations and significantly reducing computational and memory demands.

By treating the membrane time constant \( \tau \) as a form of quantization, \textit{SpikePack} adapts to various hardware configurations, balancing precision and efficiency. For instance, \( \tau = 2 \) results in uniform quantization, while \( \tau \neq 2 \) enables non-uniform quantization, allowing the model to adjust its computational footprint based on hardware constraints.

\subsubsection{Efficient Gradient Propagation}
\label{sec:spikepack_gradient}
As shown in Figure~\ref{fig:forward_backward_computation}, \textit{SpikePack} enhances computational efficiency by simplifying gradient propagation, eliminating the need for BPTT. Instead, gradients are computed directly with respect to the compressed input structure \( \mathbf{s}^{l-1}_\mathrm{zip} \), which encapsulates the entire sequence of input spikes in a single compressed representation. This method significantly reduces memory consumption and computational complexity during training by removing the requirement to unroll across time s

The gradient of the loss \( \mathcal{L} \) with respect to \( \mathbf{s}^{l-1}_\mathrm{zip} \) is computed as shown in Eq.~\eqref{eq:gradient_vg}:
\begin{equation}
    \frac{\partial \mathcal{L}}{\partial \mathbf{s}^{l-1}_\mathrm{zip}} = 
    \frac{\partial \mathcal{L}}{\partial \mathbf{s}^{l}_\mathrm{zip}}
    \frac{\partial \mathbf{s}^{l}_\mathrm{zip}}{\partial v_{g}^l} 
    \frac{\partial  v_{g}^l}{\partial \mathbf{s}^{l-1}_\mathrm{zip}}
    \approx \frac{\partial
    \mathcal{L}}{\partial \mathbf{s}^{l}_\mathrm{zip}} \frac{\mathbf{w}^l}{\theta}, \label{eq:gradient_vg}
\end{equation}
where we approximate \( \frac{\partial \lceil x \rfloor_{\tau}}{\partial x} \approx 1 \), effectively bypassing the non-differentiable quantization step. This direct gradient path allows for efficient, memory-saving training aligned with \textit{SpikePack}’s compressed and parallel computation model.

By leveraging this streamlined gradient computation, \textit{SpikePack} enables gradient propagation without costly temporal dependencies, as required in traditional SNN models. This design is inherently compatible with hardware architectures that support vectorized and parallel computation, such as SIMD instructions and systolic arrays.

    \section{Experiment}
\label{sec:exp}

We evaluate \textit{SpikePack} on tasks including image classification, object detection, and semantic segmentation, comparing its performance with state-of-the-art SNN models, neuron designs, and ANN-to-SNN conversion methods. We also conduct ablation studies to assess the impact of key parameters, demonstrating the versatility and efficiency of \textit{SpikePack} across various datasets.

\subsection{Experimental Setup}
\label{sec:experimental_setup}
We conduct experiments on both static and neuromorphic datasets to thoroughly assess our model. For image classification, we use ImageNet dataset~\cite{deng2009imagenet}; for object detection, the COCO 2017 dataset~\cite{lin2014microsoft}; and for semantic segmentation, the ADE20K dataset~\cite{zhou2017scene}. To evaluate event-based performance, we use neuromorphic datasets including CIFAR10-DVS~\cite{li2017cifar10}, DVS-Gesture~\cite{amir2017low}, and N-Caltech101~\cite{orchard2015converting}. Experiments are implemented in PyTorch and run on NVIDIA A100 GPUs, with a default membrane time constant of \( \tau = 2 \). Additional experimental details are provided in Appendix~\ref{appendix:exp}.

\subsection{Image and Neuromorphic Data Classification}
\label{sec:image_classification}

We evaluate \textit{SpikePack} on the ImageNet dataset and compare its performance with other SNN models and neuron designs.

\textbf{Comparison with Other Neuron Models}
% \label{sec:comparison_neuron_models}
We benchmark \textit{SpikePack} against several neuron models, including LIF, LIFB~\cite{shen2023exploiting}, PSN~\cite{fang2024parallel}, DSGM~\cite{shen2024exploiting}, and GLIF~\cite{yao2022glif}. For fair evaluation, experiments are conducted using the SEW-ResNet~\cite{fang2021deep} and Spikeformer~\cite{zhou2023spikformer} architectures. Model scales and time steps are adjusted to achieve comparable performance metrics across setups.

% 这个是和别的直接训练的SNN的, 不同的神经元类型的对比. 包括SEW-ResNet, 以及 Spikeformer 两种模型, 横轴是OPs, 纵轴是Accuracy, 旨在说明在相同的模型设置的时候, 我们的方法比LIF, LIFB, PSN, SDGM, GLIF 等神经元类型都好, 在相同的OPs下, 我们的方法的准确率更高, 而且参数数量不变. 
\begin{figure}[h]
    \centering
    \includegraphics[width=\linewidth]{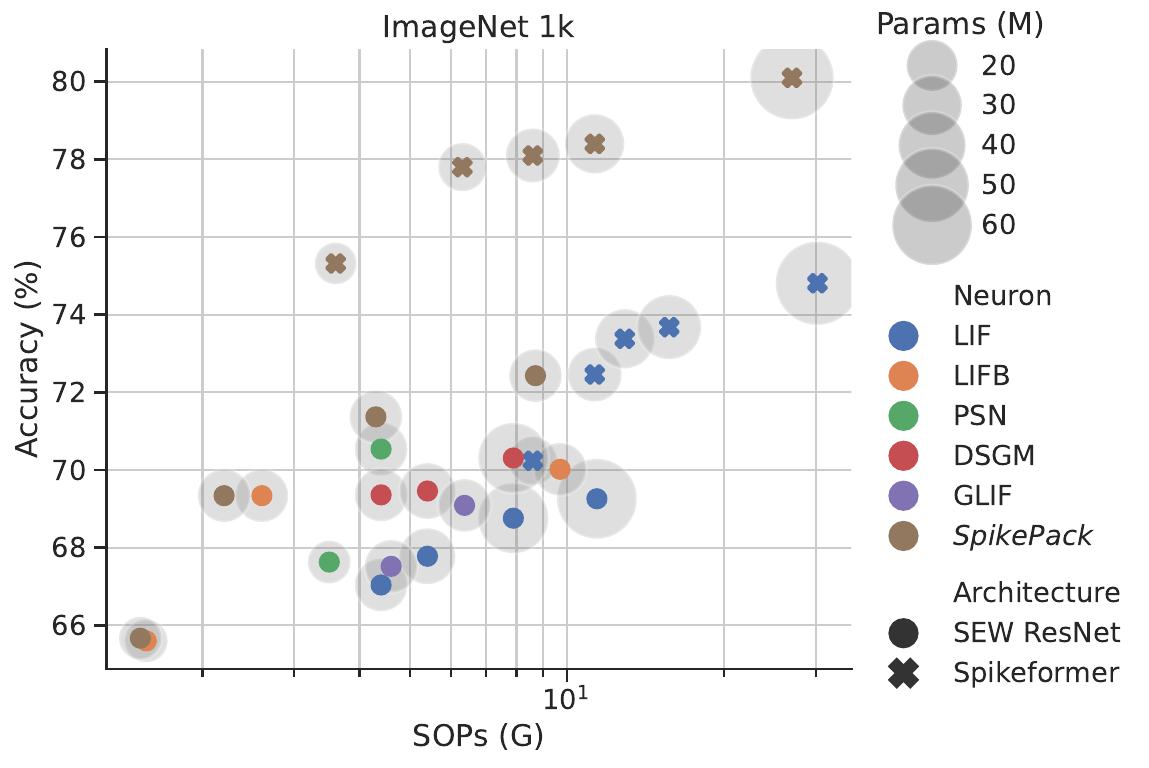}
    \caption{Comparison between \textit{SpikePack} and other neuron models at different time steps on ImageNet 1k. Our method achieves higher accuracy with lower computational cost.}
    \vspace{-2mm}
    \label{fig:neuron_comparison}
    \vspace{-2mm}
\end{figure}

% 这个是从ANN-SNN转换的和别的SNN转换方法的对比. 我们的方法只需要很短的step就能实现近似无损的精度, 还不需要额外的微调. 兼容ResNet, Transformer等各种不同的模型. 
\begin{table}[h]
    \centering
    \setlength{\tabcolsep}{1.5pt} 
    \renewcommand{\arraystretch}{1.0} 
    \caption{Comparison of ANN-to-SNN conversion methods on ImageNet 1k. We report the top-1 accuracy (\%) at different time steps (T).}
    \begin{tabular}{l|l|r|rrrr}
    \toprule
    Model     & Method &\makecell[c]{ANN}&  T=32 &  T=64 & T=128 & T=256 \\
    \midrule
    \multirow{4}{*}{ResNet-34} &    RMP~\cite{han2020deep} & -     &   -   &   -   &   -   & 55.65 \\
                              &   Opt.~\cite{deng2021optimal} & 70.64 & 33.01 & 59.52 & 67.54 & 70.06 \\
                              & Calib.~\cite{li2022efficient} & 70.95 & 64.54 & 71.12 & 73.45 & 74.61 \\
                              &    SNM~\cite{wang2022signed} & 75.66 & 55.28 & 62.72 & 65.53 & 69.31 \\
    \cline{1-7}
    ViT-B/32                  &    STA~\cite{jiang2024spatio} & 73.30 & 78.72 & 82.33 & 82.56 & 82.79 \\
    \midrule
         &  &&  T=5 &  T=6 & T=8 & T=12 \\
    \midrule
    ResNet-34 & \multirow{2}{*}{\textbf{\textit{SpikePack}}} & 80.72 & 59.32 & 74.68 & 77.81 & \textbf{77.92} \\
    ViT-B/32  &  \multirow{2}{*}{\textbf{Ours}}              & 77.92 & 57.23 & 79.26 & 80.68 & \textbf{80.72} \\
    ViT-L/14  &                                              & 88.27 & 85.59 & 88.00 & 88.22 & \textbf{88.27} \\
    \bottomrule
    \end{tabular}
    \label{tab:conversion_result}
    \vspace{-4mm}
\end{table}

Figure~\ref{fig:neuron_comparison} illustrates the efficiency of \textit{SpikePack}, showcasing its balance between accuracy and computational cost, quantified by Synaptic Operations (SOPs). SOPs measure the overall spiking activity and computational workload, defined as the product of the firing rate, operations per time step, and the number of time steps \( T \).

Compared to other neuron models, \textit{SpikePack} consistently achieves higher accuracy at similar or lower computational costs. For instance, within the Spikeformer architecture, \textit{SpikePack} matches the accuracy of competing neuron models while requiring only $1/10$ of the SOPs. Furthermore, at equivalent SOP levels, \textit{SpikePack} delivers nearly a $5\%$ improvement in accuracy, highlighting its superior information transmission and computational efficiency. This advantage persists across varying model sizes and architectures, demonstrating the robustness and scalability of \textit{SpikePack}
.

\textbf{Comparison with Efficient SNN Training Methods}
We evaluate \textit{SpikePack} against several efficient SNN training methods, including STBP-tdBN~\cite{zheng2021going}, SEW ResNet~\cite{fang2021deep}, MS ResNet~\cite{hu2024advancing}, TEBN~\cite{duan2022temporal}, TET~\cite{deng2022temporal}, OTTT~\cite{xiao2022online}, SLTT~\cite{meng2023towards}, Parallel SNN~\cite{fang2024parallel}, and T-RevSNN~\cite{huhigh2024}.

% 这张图是直接训练和别的SNN训练方法的对比, 包括参数/训练速度/显存占用/Operations/准确率
\begin{table*}[ht]
    \centering
    \setlength{\tabcolsep}{2.5pt} 
    \renewcommand{\arraystretch}{0.95} 
    \caption{Comparison of \textit{SpikePack} with state-of-the-art SNN training methods on ImageNet 1k. We report the number of parameters, time steps, training time per epoch, memory usage per image, synaptic operations (SOPs), and top-1 accuracy.}
    \vspace{-2mm}
    \begin{tabular}{lccccccc}
    \toprule
    \multirow{2}{*}{Methods}                                            & \multirow{2}{*}{Architecture} & Param & Time & Training Time & Memory & SOP & \multirow{2}{*}{Acc (\%)} \\ 
    ~ & ~ & (M) & steps & (min/ep) & (MB/img) & (G) & ~ \\ \hline
    STBP-tdBN~\cite{zheng2021going}                                                 & ResNet-34         & 21.8 & 6 & 29.6 & 186.1   & 7.1   & 63.7 \\ \cline{2-8}
    \multirow{2}*{SEW ResNet~\cite{fang2021deep}}                                   & SEW-ResNet-34     & 21.8 & 4 & 5.0  & 224.5   & 4.4   & 67.0 \\ 
                                                                                        & SEW-ResNet-50     & 25.6 & 4 & 10.0 & 596.9   & 5.4   & 67.8 \\ \cline{2-8}
    MS ResNet~\cite{hu2024advancing}                                                & MS-ResNet-34      & 21.8 & 6 & 11.2 & 267.1   & 5.7   & 69.4 \\ \cline{2-8}
    TEBN~\cite{duan2022temporal}                                                    & ResNet-34         & 21.8 & 4 & 16.3 & 260.1   & 7.1   & 64.3  \\ \cline{2-8}
    TET~\cite{deng2022temporal}                                                     & SEW-ResNet-34     & 21.8 & 4 & 12.5 & 221.0   & 4.4   & 68.0 \\ \hline
    \multirow{2}{*}{Spikformer~\cite{zhou2023spikformer}}             & Spikeformer-8-384 & 16.8 & 4 & 14.2 & 580.8   & 8.6   & 70.2 \\ 
                                                                        & Spikeformer-8-512 & 29.7 & 4 & 16.7 & 767.8   & 12.9  & 73.4 \\ \cline{2-8}
     Spike-driven                                  & Spikeformer-8-384 & 16.8 & 4 & 15.4 & 548.9   & 4.3   & 72.3 \\ 
     Transformer~\cite{yao2023spike}              & Spikeformer-8-512 & 29.7 & 4 & 18.8 & 730.0   & 5.1   & 74.6 \\ \hline
    
    OTTT~\cite{xiao2022online}                                                      & ResNet-34         & 21.8 & 6 & 24.2 & 84.1    & 6.7   & 64.2 \\ \cline{2-8}
    \multirow{2}*{SLTT~\cite{meng2023towards}}                                      & ResNet-34         & 21.8 & 6 & 18.1 & 71.7    & 6.7   & 66.2 \\
                                                                                        & ResNet-50         & 25.6 & 6 & 23.4 & 117.3   & 8.0   & 67.0 \\ \cline{2-8}
    \multirow{2}*{Parallel SNN~\cite{fang2024parallel}}                             & SEW-ResNet-18     & 11.7 & 4 & 5.8  & 138.7   & -     & 67.6 \\
                                                                                        & SEW-ResNet-34     & 21.8 & 4 & 8.3  & 179.7   & 4.4   & 70.5  \\ \cline{2-8}
    
    \multirow{2}*{T-RevSNN~\cite{huhigh2024}}                                       & ResNet-18         & 15.2 & 4 & 6.1 & 57.5     & 1.9   & 69.8 \\
                                                                                        & ResNet-18         & 29.8 & 4 & 9.1 & 85.7     & 3.1   & 73.2 \\ \hline
    \multirow{7}*{\textbf{\textit{SpikePack} (Ours)}}                                               & ResNet-18         & 11.1 & 4 & \textbf{5.8}  & \textbf{20.1}     & 1.8   & 70.6 \\
                                                                                        & ResNet-34         & 21.8 & 4 & 8.1  & 24.3     & 3.7   & 73.4 \\
                                                                                        & ResNet-50         & 25.6 & 4 & 13.5 & 53.4     & 4.1   & 78.7 \\ \cline{2-8}       
                                                                                        & Spikeformer-8-512 & 29.7 & 2 & \textbf{9.2}  & \textbf{150.5}    & 3.9   & 73.6 \\
                                                                                        & Spikeformer-8-512 & 29.7 & 4 & 9.2  & 150.5    & 7.7   & 78.4 \\
                                                                                        & Spikeformer-8-512 & 29.7 & 6 & 9.2  & 150.5    & 11.2  & 79.2 \\
                                                                                        & Spikeformer-8-512 & 29.7 & 8 & 9.2  & 150.5    & 15.1  & \textbf{80.1} \\
                                                                   
    \bottomrule
    \end{tabular}
    \label{tab:classification_comparison}
    \vspace{-4mm}
\end{table*}

As shown in Table~\ref{tab:classification_comparison}, \textit{SpikePack} outperforms these methods in accuracy while requiring less training time and memory. For instance, using ResNet-34 with $4$ time steps, \textit{SpikePack} achieves $73.4\%$ accuracy with only $8.1$ minutes per epoch and $24.3$ MB memory per image, demonstrating reduced training overhead and superior performance. Moreover, experiments with Transformer-based architectures, such as Spikeformer, confirm that \textit{SpikePack} maintains \(\mathcal{O}(1)\) time and space complexity relative to the number of time steps \( T \), achieving up to $80.1\%$ accuracy with $8$ time steps without any increase in memory or computational load as \( T \) grows.

\textbf{Comparison with ANN-to-SNN Conversion Methods} We evaluate \textit{SpikePack} for near-lossless ANN-to-SNN conversion, leveraging its compatibility with various ANN architectures as discussed in Section~\ref{sec:spikepack_parallel}. Unlike other methods, \textit{SpikePack} enables efficient conversion without requiring post-conversion training or calibration.

\begin{figure}[h]
    \vspace{-2mm}
    \centering
    \includegraphics[width=\linewidth]{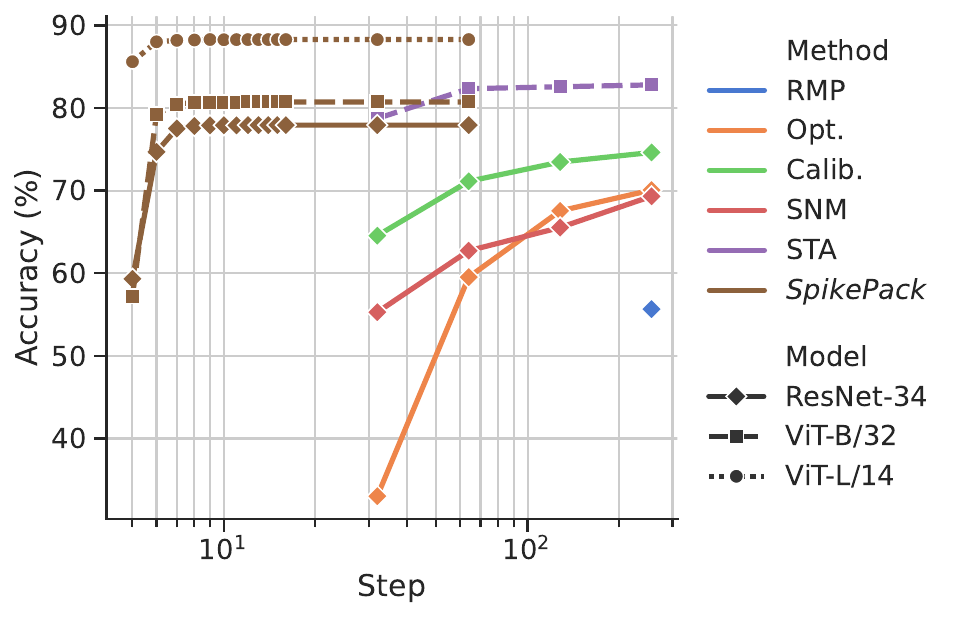}
    \vspace{-4mm}
    \caption{Comparison of ANN-to-SNN conversion methods at different time steps. \textit{SpikePack} achieves higher accuracy with fewer time steps.}
    \label{fig:conversion_result}
    \vspace{-4mm}
\end{figure}

As shown in Table~\ref{tab:conversion_result}, we compare \textit{SpikePack} with several conversion methods, including RMP~\cite{han2020deep}, Optimal (Opt.)~\cite{deng2021optimal}, Spike Calibration (Cailb.)~\cite{li2022efficient}, SNM~\cite{wang2022signed}, and Spatio-Temporal Approximation (STA)~\cite{jiang2024spatio}. \textit{SpikePack} achieves high accuracy with as few as 6 time steps and near-lossless performance at 8 time steps—less than $1/10$ of the steps required by other methods. For example, \textit{SpikePack} achieves $77.92\%$ accuracy with ResNet-34 and $88.27\%$ with ViT-L/14, closely matching ANN performance. Figure~\ref{fig:conversion_result} further highlights \textit{SpikePack}'s ability to maintain high accuracy across different architectures with minimal time steps compared to competing methods.

\textbf{Results on Neuromorphic Datasets} We evaluate \textit{SpikePack} on neuromorphic datasets including CIFAR10-DVS~\cite{li2017cifar10}, DVS-Gesture~\cite{amir2017low}, and N-Caltech101~\cite{orchard2015converting}. Table~\ref{tab:neuromorphic_result} reports the accuracy at different time steps.

\begin{table}[h]
    \centering
    \setlength{\tabcolsep}{1.8pt} 
    \renewcommand{\arraystretch}{1.} 
    \caption{Classification accuracy on neuromorphic datasets at different time steps \( T \).}
    \vspace{-2mm}
    \begin{tabular}{l|cccccc}
        \toprule
        Dataset &  T=4  &   T=6  &   T=8  &   T=10  &  T=12  &  T=16 \\
        \midrule
        CIFAR10-DVS~\cite{li2017cifar10}       &  68.1 &   76.4 &   80.7 &   82.4  &  83.7  &  84.6 \\ 
        DVS-Gesture~\cite{amir2017low}           &  94.6 &   95.5 &   96.7 &   97.4  &  96.7  &  97.4 \\ 
        N-Caltech101~\cite{orchard2015converting} &  76.3 &   80.0 &   81.7 &   82.2  &  82.5  &  82.6 \\ 
        \bottomrule
    \end{tabular}
    \label{tab:neuromorphic_result}
    \vspace{-4mm}
\end{table}

% Our method achieves competitive accuracy on neuromorphic datasets, demonstrating its adaptability to event-based data. Unlike static datasets, neuromorphic datasets generally require a higher number of time steps \( T \) for optimal performance—a trend consistent with other literature~\cite{shen2024conventional}. This result confirms \textit{SpikePack}'s versatility, as it effectively handles diverse data types. 
\textit{SpikePack} achieves competitive accuracy on neuromorphic datasets, demonstrating adaptability to event-based data. Unlike static datasets, neuromorphic datasets generally require a higher number of time steps \( T \) for optimal performance—a trend consistent with prior studies~\cite{shen2024conventional}. These results confirm \textit{SpikePack}'s versatility in effectively handling both static and event-driven data.

% Notably, \textit{SpikePack} reaches 84.6\% accuracy on CIFAR10-DVS with 16 time steps.

\subsection{Object Detection}
\label{sec:object_detection}

We evaluate \textit{SpikePack} on the COCO 2017 validation dataset~\cite{lin2014microsoft}, leveraging RTMDet~\cite{lyu2022rtmdet} and DINO~\cite{zhang2022dino} as our base architectures. To highlight the effectiveness of our approach, we compare it against other SNN-based object detection models, including Spiking-YOLO~\cite{kim2020spiking}, Bayesian Optimization~\cite{kim2020towards}, Spike Calibration~\cite{li2022efficient}, EMS-YOLO~\cite{su2023deep}, Meta-SpikeFormer~\cite{yaospiked}, and SpikeYOLO~\cite{luo2024integer}.

Table~\ref{tab:coco_result} summarizes the evaluation results. Our approach demonstrates superior mean Average Precision (mAP) while requiring fewer time steps and reducing computational cost. For instance, with DINO-r50 and just $6$ time steps, \textit{SpikePack} achieves an impressive $48.5\%$ mAP@50:95, surpassing previous methods with a substantial reduction in computational overhead.

% 这个是在目标检测的方面的实验, 看表说话就行了. 

Our method demonstrates competitive performance with significantly fewer parameters and computational cost. For instance, using RTMDet-m with only $67.2$G SOPs, we achieve $49.1\%$ mAP@50:95 at $8$ time steps.

\subsection{Semantic Segmentation}
\label{sec:semantic_segmentation}

\begin{table}[h]
    \caption{Performance of object detection on COCO 2017 validation set~\cite{lin2014microsoft}. We report the number of parameters, computational cost (SOPs), time steps, and mean Average Precision (mAP).}
    \vspace{-2mm}
    \label{tab:coco_result}
    % \footnotesize
    \centering
    \setlength{\tabcolsep}{1pt}
    \renewcommand{\arraystretch}{1.}
    \begin{tabular}{lcccccc} 
        \toprule
  
        \multirow{2}{*}{Model}                          & Param & SOPs         & \multirow{2}{*}{Step}&  mAP@  & mAP@ \\ 
                                                        & (M)   & (G)         &         &  50(\%)  & 50:95(\%) \\
        \midrule
        Spiking-YOLO~\cite{kim2020spiking}              &  10.2 & -   & 3500  & -    & 25.7       \\
        Bayesian Optim~\cite{kim2020towards}            &  10.2 & -   & 5000  & -    & 25.9       \\
        Spike Calib~\cite{li2022efficient}              &  17.1 & -   & 512   & 45.4 & -           \\
        \cline{1-6}  
        EMS-YOLO\cite{su2023deep}                       &  26.9  & 32.2   & 4 & 50.1 & 30.1 \\
        \cline{1-6}  
        Meta-SpikeFormer                                &  34.9  & 55.0   & 1 & 44.0 & - \\
        (MaskRCNN)~\cite{yaospiked}                     &  75.0  & 156.4  & 1 & 51.2 & - \\
        \cline{1-6}        
        Meta-SpikeFormer                                &  16.8  & 38.7   & 1 & 45.0 & - \\
        (YOLO)~\cite{yaospiked}                     &  16.8  & 78.6   & 4 & 50.3 & - \\
        \cline{1-6}                   
        \multirow{3}{*}{SpikeYOLO~\cite{luo2024integer}}&  23.1  & 38.6   & 4 & 62.3 & 45.5 \\ 
                                                        &  48.1  & 76.1   & 4 & 64.6 & 47.4  \\ 
                                                        &  68.8  & 93.6   & 4 & 66.2 & 48.9 \\   
        \cline{1-6}       
        \multirow{3}{*}{\makecell[l]{\textbf{\textit{SpikePack}} (Ours)\\w/ RTMDet-tiny~\cite{lyu2022rtmdet}}}  &  \textbf{4.8}   & 8.53   & 6  & 55.7 & 39.0 \\ 
                                                                    &  4.8   & 11.3   & 8  & 57.8 & 40.9 \\ 
                                                                    &  4.8   & 14.1   & 10 & 57.9 & 41.1 \\ 
        \cline{2-6}   
        \multirow{3}{*}{\makecell[l]{\textbf{\textit{SpikePack}} (Ours)\\w/ RTMDet-m~\cite{lyu2022rtmdet}}}     &  24.7  & 51.8   & 6 & 50.1 & 48.5 \\ 
                                                                    &  24.7  & 67.2     & 8 & 61.7 & 49.1 \\
                                                                    &  24.7  & 86.3     & 10 & 61.9 & 49.4 \\
        \cline{2-6} 
        \multirow{3}{*}{\makecell[l]{\textbf{\textit{SpikePack}} (Ours)\\w/ DINO-r50~\cite{zhang2022dino}}}     & 47.7   & 276   & 6 & 66.0 & 48.5 \\ 
                                                                    & 47.7   & 359   & 8 & 66.7 & 50.0 \\ 
                                                                    & 47.7   & 447   & 10 & \textbf{67.9} & \textbf{50.1} \\ 
        \bottomrule
    \end{tabular}  
\end{table}

Table~\ref{tab:ade20k_result} presents the segmentation results on ADE20K, showcasing \textit{SpikePack}'s strong performance in dense prediction tasks like semantic segmentation. With Segformer-b2 and $10$ time steps, \textit{SpikePack} achieves $45.6\%$ mIoU, outperforming prior methods while significantly reducing computational cost. These results emphasize the method's efficiency and scalability for challenging benchmarks.

% 这个是在语义分割的实验, 看表说话就行了.
\begin{table}[h]
    \caption{Performance of semantic segmentation on ADE20K~\cite{zhou2017scene}. We report the number of parameters, computational cost (SOPs), simulation time steps, and mIoU (\%).}
    \vspace{-2mm}
    \label{tab:ade20k_result}
    % \footnotesize
    \centering
    \setlength{\tabcolsep}{4pt} 
    \renewcommand{\arraystretch}{.9} 
    \begin{tabular}{lcccc}  
        \toprule
        \multirow{2}{*}{Model}                          & Param & SOPs         & \multirow{2}{*}{Step}&  MIoU(\%)  \\ 
                                                        & (M)   & (G)         &         &  50(\%)   \\
        \midrule
        \multirow{4}{*}{Meta-SpikeFormer~\cite{yaospiked}}&  16.5   & 24.6   & 1 & 32.3  \\
                                                        &  16.5     & 98.2   & 4 & 33.6  \\
        \cline{2-5}                           
                                                        &  58.9     & 51.7   & 1 & 34.8  \\
                                                        &  58.9     & 204.1  & 4 & 35.3  \\
        \cline{1-5}                   
        \multirow{3}{*}{\makecell[l]{\textbf{\textit{SpikePack}} (Ours)\\w/ FCN-r50}~\cite{long2015fully}}     
                                                        & 47.2   & 256.8   & 6 & 34.1 \\ 
                                                        & 47.2   & 384.4   & 8 & 35.3  \\ 
                                                        & 47.2   & 476.7   & 10 & 35.9  \\ 
        \cline{1-5}   
        \multirow{3}{*}{\makecell[l]{\textbf{\textit{SpikePack}} (Ours)\\w/ Segformer-b0~\cite{xie2021segformer}}}   
                                                        &  \textbf{3.75}  &  \textbf{39.3}   & 6  & 35.3 \\ 
                                                        &  3.75  &  51.7   & 8  & 36.9 \\
                                                        &  3.75  &  63.9   & 10 & 37.4 \\     
        \cline{2-5}   
        \multirow{3}{*}{\makecell[l]{\textbf{\textit{SpikePack}} (Ours)\\w/ Segformer-b2~\cite{xie2021segformer}}}   
                                                       &  24.8  &  83.6    & 6  & 42.8 \\ 
                                                        &  24.8  &  111.5   & 8  & 44.1 \\
                                                        &  24.8  &  138.7   & 10 & \textbf{45.6} \\
        \bottomrule
    \end{tabular}  
    \vspace{-4mm}
\end{table}

\subsection{Ablation Studies}
\label{sec:ablation_studies}

\begin{figure}[h]
    \centering
    \includegraphics[width=\linewidth]{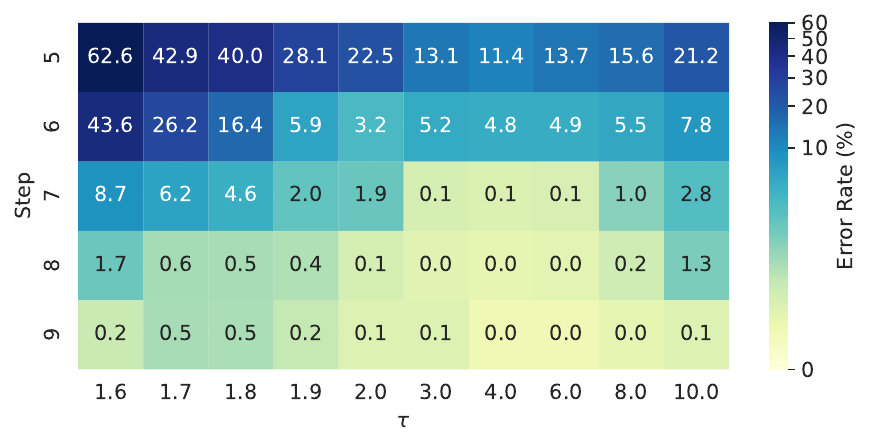}
    \vspace{-4mm}
    \caption{Ablation study on the effect of membrane time constant $\tau$ on ImageNet classification error rate (\%) using ResNet-34 with \textit{SpikePack}.}
    \label{fig:ablation_tau}
    \vspace{-4mm}
\end{figure}

We conduct ablation studies to analyze the impact of the membrane time constant $\tau$ on \textit{SpikePack}'s performance, evaluated on ImageNet with ResNet-34 (Figure~\ref{fig:ablation_tau}). While $\tau=4.0$ achieves the best performance by aligning with activation distributions, $\tau=2.0$ is more efficient on GPGPUs, avoiding exponentiation. As a result, $\tau=2.0$ has been the preferred choice in prior implementations.

\subsection{Hardware Compatibility}

\begin{table}[h]
    \centering
    \setlength{\tabcolsep}{1.5pt}
    \renewcommand{\arraystretch}{.9}
    \caption{Latency and Energy Comparison between SpikePack and LIF neuron}
    \vspace{-2mm}
\begin{tabular}{c|cc|cc}
\hline
      & \multicolumn{2}{c|}{\textit{\textbf{SpikePack}}}           & \multicolumn{2}{c}{LIF}                  \\ \hline
       & \multicolumn{1}{c|}{ResNet-34} & ResNet-50 & \multicolumn{1}{c|}{ResNet-34} & ResNet-50 \\ \hline
Latency & \multicolumn{1}{c|}{23ms}       & 24.1 ms     & \multicolumn{1}{c|}{29.1 ms}     & 34.7 ms     \\ \hline
Energy  & \multicolumn{1}{c|}{18.6mJ}     & 19.4 mJ     & \multicolumn{1}{c|}{23.8 mJ}     & 28.4 mJ     \\ \hline
\end{tabular}
\vspace{-2mm}
\end{table}

\textbf{Compatibility with Neuromorphic Processors}  
As a spiking neuron, the \textit{SpikePack} neuron adheres to the binary nature of spiking neurons, generating discrete spike sequences that are fully compatible with existing neuromorphic processors. With higher sparsity, \textit{SpikePack} achieves superior speedup and reduced energy consumption compared to traditional LIF neurons. To validate this, we conduct hardware experiments comparing \textit{SpikePack} and LIF neurons under identical conditions. We design a neuromorphic processor that processes binary spike inputs and synaptic weights, filtering zero elements in the spike tensor to ensure only active spikes contribute to computation. The processor includes 64 processing elements for synaptic current accumulation, 16 neuron dynamic units, and a 16-input spike detector that converts spike sequences into addresses for synaptic weight fetching.
We implement the processor on xczu3eg FPGA chip running at 300MHz. The detailed implementation is shown in Appendix~\ref{appendix:hardware}.
Using cycle-accurate simulation, we evaluate inference latency for ResNet-34 and ResNet-50 models. The results demonstrate that \textit{SpikePack} achieves significantly lower latency and power consumption than traditional LIF neurons, highlighting its efficiency on neuromorphic hardware.

\textbf{Compatibility with Parallel Computing Processors}  
Modern parallel computing processors, such as GPGPUs, NPUs, and SIMD-enabled CPUs, excel at matrix multiplication—a critical operation in neural network acceleration. The \textit{SpikePack} neuron efficiently utilizes these architectures by maintaining \(\mathcal{O}(1)\) computation and memory usage across time steps. In contrast, traditional LIF neurons require \(\mathcal{O}(T)\) duplication of computational workload and additional storage for membrane potentials, resulting in increased computational overhead. 
    \subsection{Discussion}

Our experimental results demonstrate that \textit{SpikePack} significantly improves information flow and computational efficiency in SNNs. By minimizing information loss during spike transmission and supporting efficient parallel computation, \textit{SpikePack} delivers superior performance across diverse tasks and datasets. Additionally, its seamless compatibility with standard ANN architectures enables near-lossless ANN-to-SNN conversion, allowing SNNs to benefit from the latest advancements in ANN models and training techniques.

    { \small \bibliographystyle{ieeenat_fullname} \bibliography{main} }

    \clearpage
\setcounter{page}{1}
\maketitlesupplementary

\appendix
\section{Mutual Information Analysis of \textit{SpikePack}}
\label{appendix:mutual_information}

In this appendix, we provide a formal analysis of the mutual information properties of the proposed \textit{SpikePack} neuron model, comparing it with the traditional Leaky Integrate-and-Fire (LIF) neuron model. This analysis aims to show that \textit{SpikePack} neurons retain more information between pre-synaptic inputs and post-synaptic outputs, thereby reducing information loss during spike transmission.

\subsection{Problem Statement}

Consider a spiking neuron receiving binary input spikes over \( T \) time steps from \( N \) pre-synaptic neurons. Let \( \mathbf{S}^{l} \in \{0,1\}^{N \times T} \) denote the input spike matrix, where each element \( s^{l}_{n, t} \) represents the spike from the \( n \)-th neuron at time step \( t \). Each spike \( s^{l}_{n, t} \) is assumed to be an independent Bernoulli random variable with parameter \( p \), i.e., \( s^{l}_{n, t} \sim \text{Bernoulli}(p) \). The synaptic weights are represented by \( \mathbf{w} \in \mathbb{R}^N \), where each weight \( w_n \) is drawn independently from a Gaussian distribution \( \mathcal{N}(0, \sigma^{2}) \).

Our objective is to compute and compare the mutual information \( I(\mathbf{S}^{l}; \mathbf{s}^{l}) \) between input and output spikes for both \textit{SpikePack} and LIF neurons.

\subsection{Mutual Information in \textit{SpikePack} Neurons}

\paragraph{Accumulated Membrane Potential}

In the \textit{SpikePack} neuron, the accumulated membrane potential \( v_{g}^l \) is defined as:
\begin{equation}
    v_{g}^l = \mathbf{w}^{\top} \mathbf{S}^{l-1} \mathbf{q},
    \label{eq:appendix_v0}
\end{equation}
where \( \mathbf{q} = [\tau^{T - 1}, \tau^{T - 2}, \dots, \tau^{0}]^{\top} \) incorporates the effect of leakage across time steps.

\paragraph{Distribution of $v_{g}^l$}

Given that the input spikes are independent Bernoulli random variables and the weights are independent Gaussian random variables, the accumulated membrane potential \( v_{g}^l \) is a sum of independent random variables. By the Central Limit Theorem, \( v_{g}^l \) approximates a Gaussian distribution when \( N \) is large.

\textbf{Mean of \( v_{g}^l \):}
\begin{equation}
    \mu_{v_{g}^l} = \mathbb{E}[v_{g}^l] = \sum_{n=1}^{N} \mathbb{E}[w_n] \sum_{t=1}^{T} \mathbb{E}[s^{l}_{n, t}] q_t = 0,
\end{equation}
since \( \mathbb{E}[w_n] = 0 \).

\textbf{Variance of \( v_{g}^l \):}
\begin{equation}
    \sigma_{v_{g}^l}^{2} = \mathbb{E}[{v_{g}^l}^{2}] = \sigma^{2} N p (1 - p) \left( \sum_{t=1}^{T} q_t \right)^{2},
    \label{eq:appendix_sigma_v0}
\end{equation}
where \( q_t = \tau^{t - 1} \).

\paragraph{Differential Entropy of $v_{g}^l$}

Since \( v_{g}^l \) is approximately Gaussian with variance \( \sigma_{v_{g}^l}^{2} \), its differential entropy is:
\begin{equation}
    h(v_{g}^l) = \frac{1}{2} \log_{2}(2\pi e \sigma_{v_{g}^l}^{2}).
    \label{eq:appendix_h_v0}
\end{equation}

\paragraph{Conditional Entropy $h(v_{g}^l | \mathbf{s}^{l})$}

The \textit{SpikePack} neuron generates output spikes \( \mathbf{s}^{l} \) by quantizing the continuous membrane potential \( v_{g}^l \) with a quantization step size \( \theta \). This process introduces quantization noise, as \( v_{g}^l \) is mapped to the nearest discrete level defined by \( \theta \). Following the approach in \cite{widrow2008quantization}, we assume that this quantization noise is uniformly distributed over \( \left[-\frac{\theta}{2}, \frac{\theta}{2}\right] \). This assumption is valid when the quantization step size \( \theta \) is relatively small compared to the variance of \( v_{g}^l \), and the signal \( v_{g}^l \) is approximately Gaussian and sufficiently random. 

Given that the quantization noise \( q \) is uniformly distributed over \( \left[-\frac{\theta}{2}, \frac{\theta}{2}\right] \), the probability density function of \( q \) is:
\begin{equation}
f(q) = 
\begin{cases}
    \frac{1}{\theta} & \text{for } -\frac{\theta}{2} \leq q \leq \frac{\theta}{2}, \\
    0 & \text{otherwise}.
\end{cases}
\end{equation}

The conditional entropy \( h(v_{g}^l | \mathbf{s}^{l}) \) represents the uncertainty introduced by quantizing \( v_{g}^l \) and is equal to the entropy of the quantization noise \( q \) over the interval \( \left[-\frac{\theta}{2}, \frac{\theta}{2}\right] \). The entropy of a continuous uniform distribution is calculated as:
\begin{equation}
h(v_{g}^l | \mathbf{s}^{l}) = \int_{-\theta/2}^{\theta/2} -f(q) \log_2(f(q)) \, dq.
\end{equation}

Substituting \( f(q) = \frac{1}{\theta} \), we get:
\begin{equation}
h(v_{g}^l | \mathbf{s}^{l}) = \log_2(\theta).
\end{equation}

To refine this result, we apply a correction factor for the entropy of the uniform distribution, considering its variance. For a uniform distribution over \( \left[-\frac{\theta}{2}, \frac{\theta}{2}\right] \), the variance is \( \text{Var}(q) = \frac{\theta^2}{12} \)~\cite{widrow2008quantization}., so the standard deviation is \( \frac{\theta}{\sqrt{12}} \). Thus, the correction term \( \log_2(\sqrt{12}) \) accounts for the spread of the distribution:
\begin{equation}
h(v_{g}^l | \mathbf{s}^{l}) = \log_2(\theta) - \log_2(\sqrt{12}).
\end{equation}

This refined expression for the conditional entropy \( h(v_{g}^l | \mathbf{s}^{l}) \) accurately reflects the quantization effects within the \textit{SpikePack} neuron model.

\paragraph{Mutual Information Calculation}

The mutual information between \( v_{g}^l \) and \( \mathbf{s}^{l} \) is:
\begin{equation}
    I(v_{g}^l; \mathbf{s}^{l}) = h(v_{g}^l) - h(v_{g}^l | \mathbf{s}^{l}) = \frac{1}{2} \log_{2}\left( \frac{12 \sigma_{v_{g}^l}^{2}}{\theta^{2}} \right).
    \label{eq:appendix_I_v0_s_out}
\end{equation}

Since \( \mathbf{s}^{l} \) is a deterministic function of \( v_{g}^l \), we have:
\begin{equation}
    I(\mathbf{S}^{l}; \mathbf{s}^{l}) = I(v_{g}^l; \mathbf{s}^{l}).
    \label{eq:appendix_I_S_in_s_out}
\end{equation}

Thus, the mutual information for the \textit{SpikePack} neuron is:
\begin{equation}
    I_{\mathrm{SP}} = \frac{1}{2} \log_{2}\left( \frac{12 \sigma_{v_{g}^l}^{2}}{\theta^{2}} \right).
    \label{eq:appendix_I_SP}
\end{equation}

\subsection{Mutual Information in LIF Neurons}

\paragraph{Approximated Membrane Potential}

In the LIF neuron, the recursive membrane potential update complicates a direct calculation of mutual information. We approximate the membrane potential at time \( t \) as:
\begin{equation}
    v'_{t} = \mathbf{w}^{\top} \mathbf{s}^{l}_{t},
    \label{eq:appendix_v_prime_t}
\end{equation}
ignoring temporal dependencies and leakage.

\paragraph{Distribution of $v'_{t}$}

Each \( v'_{t} \) is approximately Gaussian with mean zero and variance:
\begin{equation}
    \sigma_{v'_{t}}^{2} = \sigma^{2} N p (1 - p).
    \label{eq:appendix_sigma_v_prime_t}
\end{equation}

\paragraph{Probability of Spiking and Entropy of Output Spikes}

The probability of an output spike at time \( t \) is:
\begin{equation}
    P(s'_{\mathrm{out}, t} = 1) = Q\left( \frac{\theta}{\sigma_{v'_{t}}} \right),
    \label{eq:appendix_P_s_out_prime_t}
\end{equation}
where \( Q(\cdot) \) is the Q-function. Using this probability, the entropy of the output spike at each time step is:
\begin{equation}
    \begin{split}
        H(s'_{\mathrm{out}, t}) = & -P(s'_{\mathrm{out}, t} = 1) \log_{2} P(s'_{\mathrm{out}, t} = 1) \\
        & - P(s'_{\mathrm{out}, t} = 0) \log_{2} P(s'_{\mathrm{out}, t} = 0).
    \end{split}
    \label{eq:appendix_H_s_out_prime_t}
    \end{equation}

\paragraph{Upper Bound on Mutual Information}

Assuming independence across time steps, the total mutual information is bounded by:
\begin{equation}
    I_{\mathrm{LIF}} = I(\mathbf{S}^{l}; \mathbf{s}^{l}) \leq \sum_{t=1}^{T} H(s'_{\mathrm{out}, t}).
    \label{eq:appendix_I_LIF}
\end{equation}

\subsection{Comparative Analysis and Numerical Estimation}

\paragraph{Parameter Settings}

We use the following parameters for both theoretical and numerical estimation:
\begin{itemize}
    \item Number of pre-synaptic neurons: \( N = 16 \)
    \item Number of time steps: \( T = 16 \)
    \item Weight variance: \( \sigma^{2} = 1 \)
    \item Input spike probability: \( p = 0.5 \)
    \item Membrane time constant: \( \tau = 2 \)
\end{itemize}

\paragraph{\textit{SpikePack} Mutual Information Calculation}

Compute \( \sigma_{v_{g}^l}^{2} \) using Eq.~\eqref{eq:appendix_sigma_v0}:
\begin{equation}
    \sigma_{v_{g}^l}^{2} = 4 \left( 2^{16} - 1 \right)^{2}.
\end{equation}

Substitute \( \sigma_{v_{g}^l}^{2} \) and \( \theta = \frac{6 \sigma_{v_{g}^l}}{2^{T}} \) into Eq.~\eqref{eq:appendix_I_SP}:
\begin{equation}
    I_{\mathrm{SP}} \approx 15.21 \text{ bits}.
\end{equation}

\paragraph{LIF Neuron Mutual Information Calculation}

For the LIF neuron, \( \sigma_{v'_{t}}^{2} = 4 \) and \( P(s'_{\mathrm{out}, t} = 1) = Q(0.5) \approx 0.3085 \). Using Eq.~\eqref{eq:appendix_H_s_out_prime_t}, each time step contributes approximately \( H(s'_{\mathrm{out}, t}) \approx 0.881 \) bits, leading to:
\begin{equation}
    I_{\mathrm{LIF}} \leq 16 \times 0.881 = 14.096 \text{ bits}.
\end{equation}

\paragraph{Comparison and Interpretation}

The mutual information estimates indicate that:
\begin{itemize}
    \item \textit{SpikePack} achieves \( I_{\mathrm{SP}} \approx 15.21 \) bits.
    \item LIF Neuron achieves \( I_{\mathrm{LIF}} \leq 14.096 \) bits.
\end{itemize}

This demonstrates that \textit{SpikePack} retains more information, validating the theoretical analysis.

\subsection{Empirical Validation}

To validate our theoretical findings, we conducted Monte Carlo simulations to estimate \( I(\mathbf{S}^{l}; \mathbf{s}^{l}) \) for both neuron models under various configurations of \( N \) and \( T \). The results, depicted in Figure~\ref{fig:mi}, Section~\ref{sec:spikepack_neuron}, confirm that \textit{SpikePack} neurons consistently achieve higher mutual information than LIF neurons across different settings, reinforcing the conclusion that \textit{SpikePack} effectively reduces information loss during spike transmission.

% \subsection{Conclusion}

This analysis shows that the \textit{SpikePack} neuron model achieves higher mutual information between input and output spikes than the LIF neuron model. By aggregating information across time steps before spike generation, \textit{SpikePack} reduces information loss and enhances transmission efficiency, supporting more effective information flow in SNNs.

\section{Experimental Details}
\label{appendix:exp}

In this section, we provide a comprehensive description of the datasets, model architectures, and hyperparameter settings used in our experiments. This includes details on both static image and neuromorphic datasets, as well as specific training configurations for each task.

\subsection{Datasets}
We evaluate \textit{SpikePack} on both static and neuromorphic datasets to assess its performance across a range of visual tasks.

\paragraph{Static Datasets}

\begin{itemize}
    \item {ImageNet}~\cite{deng2009imagenet}: A large-scale image dataset containing over one million images categorized into 1,000 classes. This dataset provides diverse and complex visual content, which is crucial for evaluating classification performance on high-resolution images. For ImageNet, we resize images to $224 \times 224$.
    \item {COCO 2017}~\cite{lin2014microsoft}: A widely-used benchmark for object detection, containing 118,000 training images and 5,000 validation images with 80 object categories. We use this dataset to test \textit{SpikePack} on object detection tasks.
    \item {ADE20K}~\cite{zhou2017scene}: A semantic segmentation dataset with over 20,000 training images covering 150 classes. ADE20K provides a challenging setup for testing dense pixel-wise prediction tasks, such as segmentation.
\end{itemize}

\paragraph{Neuromorphic Datasets}
\begin{itemize}
    \item {CIFAR10-DVS}~\cite{li2017cifar10}: A neuromorphic adaptation of CIFAR-10, generated using a Dynamic Vision Sensor (DVS) to capture asynchronous event streams. The dataset consists of 10 classes, matching the original CIFAR-10 categories, with each sample transformed into a sequence of events.
    \item {DVS-Gesture}~\cite{amir2017low}: A dataset designed for gesture recognition, containing hand gestures captured from different individuals under varying lighting conditions. The dataset offers dynamic and complex temporal patterns that challenge spiking models.
    \item {N-Caltech101}~\cite{orchard2015converting}: This dataset is a neuromorphic version of the Caltech101 object classification dataset, generated through a DVS camera that records event-based sequences for 101 object categories.
\end{itemize}

\subsection{Hyperparameters and Configuration}

For our experiments, we evaluate \textit{SpikePack} in two settings: direct training and ANN-to-SNN conversion. 

In the direct training setup, we adhere to the settings used by Zhou et al.~\cite{zhou2022spikformer} for comparability and consistency. For ImageNet datasets, the input resolution is set to $224 \times 224$, unless otherwise noted in the main text. Neuromorphic datasets are resized to $48 \times 48$ to streamline computational costs. Batch size is dynamically adjusted according to the specific model architecture, maximizing memory usage without exceeding 40 GB of GPU memory. We employ native Automatic Mixed Precision (AMP) for all training processes to balance computational efficiency and memory usage. The initial learning rate is set to $0.001$, and models are trained for 300 epochs unless otherwise specified. The membrane time constant $\tau$ is set to 2 by default, and threshold $\theta$ is dynamically adjusted as $\theta = {T}/{2^T}$, where $T$ is the number of time steps. This approach progressively reduces the threshold over time, creating finer divisions of the input signal, which improves information transmission over longer sequences.

For the ANN-to-SNN conversion experiments, we first calibrate $\theta$ by selecting 10\% of the training data. This subset is used to set $\theta$ in a way that minimizes the risk of overflow during inference. For evaluation, this threshold $\theta$ remains fixed to ensure stable performance across the entire test set. During conversion, $\theta$ is allocated independently for each channel, enabling fine-grained control over the activation dynamics and improving the robustness of the converted SNN model.

The computation of Synaptic Operations (SOP) follows the same procedure as Zhou et al.~\cite{zhou2022spikformer}, where SOP is defined as \( \text{SOP} = \text{fr} \times \text{FLOPs} \times T \). Here, $\text{fr}$ represents the firing rate, or the proportion of spikes generated over the total possible activations, allowing for a direct comparison of energy efficiency across models with different firing dynamics and time steps.

For object detection and semantic segmentation tasks, we apply the ANN-to-SNN conversion approach, given the high accuracy already achieved through this method. This setup maintains the accuracy benefits of the ANN models while allowing efficient deployment in SNN form, leveraging the sparsity and reduced computational costs enabled by \textit{SpikePack}.

\section{Hardware Experiments}
\label{appendix:hardware}

\begin{figure}[h]
    \centering
    \includegraphics[width=1.0\linewidth]{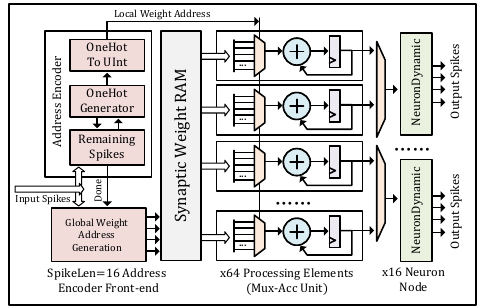}
    \caption{Hardware architecture of the neuromorphic-like processor demo customized for \textit{SpikePack} or LIF neuron.}
    \label{fig:arch}
\end{figure}

To evaluate the performance of \textit{SpikePack} neurons in comparison to traditional Leaky LIF neurons on hardware, we designed a customized digital processor resembling a neuromorphic chip. This processor processes binary spike inputs and synaptic weights, performing event-driven accumulation of synaptic currents. The architecture comprises three primary components: (1) a spike address encoder, which encodes pre-synaptic input spikes to addresses for retrieving the corresponding synaptic weights, (2) an array of processing elements (PEs) with vectorized multiplex-accumulate logic, and (3) parallel neuron node logic responsible for generating output spikes, as depicted in Figure.\ref{fig:arch}. The customized architecture builds upon and extends the FireFly-S\cite{li2024fireflys} implementation.

\begin{figure}[h]
    \centering
    \includegraphics[width=1.0\linewidth]{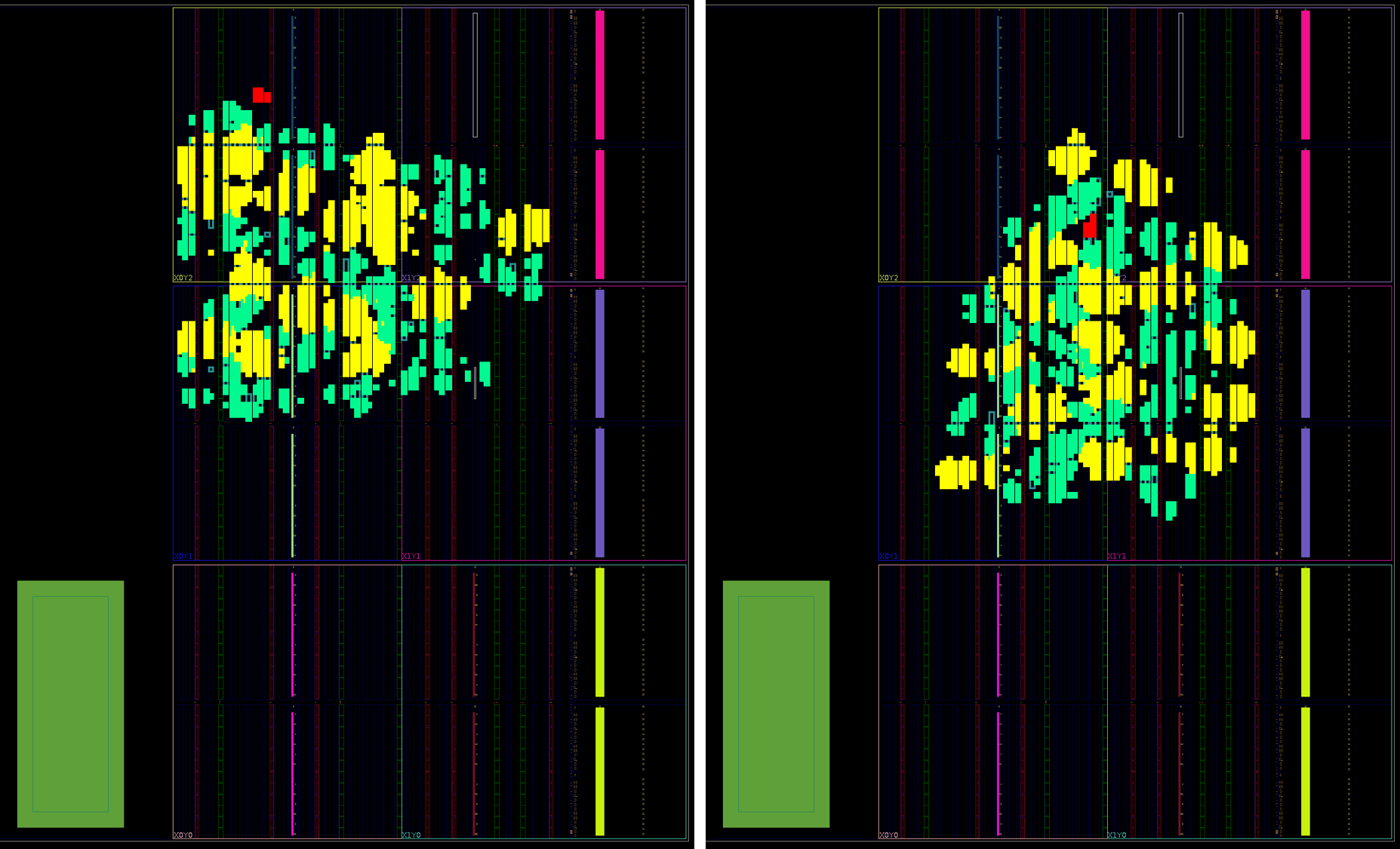}
    \caption{Hardware implementation device map of neuromorphic-like processor for \textit{SpikePack} (right) and LIF (left) neuron on xczu3eg FPGA. Color green area indicates the logic of the processing elements, color yellow indicates the logic of the \textit{SpikePack} or the LIF neuron and color red indicates the logic of the address encoder.}
    \label{fig:devicemap}
\end{figure}

The processor was tailored for both \textit{SpikePack} and LIF neurons, using a shared encoder and PE logic but differing in neuron implementation logic. Table.\ref{tab:resource} presents the resource consumption of the designs implemented on an XCZU3EG FPGA. In this analysis, we focus on logic resource utilization, excluding on-chip RAM, as synaptic weight data is directly fed from the simulation environment. The device mapping results of two implmentations are shown in Figure.\ref{fig:devicemap}.

The SpikePack implementation demonstrates a slight reduction in resource consumption compared to the traditional LIF neuron. This efficiency arises from the elimination of the need to store long-term membrane potential in hardware. Additionally, the \textit{SpikePack} implementation consumes less power, operating at 0.808 W compared to 0.816 W for the LIF implementation, both running at 300 MHz.

% \usepackage{multirow}
\begin{table}[h]
\centering
\caption{Resource consumption breakdown of customized neuromorphic-like processor for \textit{SpikePack} and LIF neuron.}
\begin{tabular}{c|c|c|c|c}
\hline
                           &        & LUTs & FFs  & CARRY8s \\ \hline
\multirow{4}{*}{\textit{SpikePack}} & Total  & 9496 & 1042 & 704     \\ \cline{2-5} 
                           & Encode & 46   & 18   & 0       \\ \cline{2-5} 
                           & PE     & 4673 & 1024 & 256     \\ \cline{2-5} 
                           & Node   & 4521 & 0    & 448     \\ \hline
\multirow{4}{*}{LIF}       & Total  & 9850 & 1302 & 768     \\ \cline{2-5} 
                           & Encode & 46   & 18   & 0       \\ \cline{2-5} 
                           & PE     & 4673 & 1024 & 256     \\ \cline{2-5} 
                           & Node   & 4875 & 260  & 512     \\ \hline
\end{tabular}
    \label{tab:resource}
\end{table}

The ResNet inference latency was measured using a cycle-accurate simulator of the proposed hardware architecture. The spike encoder effectively eliminates redundant spikes, resulting in an inference latency that is strongly correlated with the sparsity level of the spike input. As \textit{SpikePack} inherently produces a more sparse spike output pattern, it achieves lower inference latency and energy per inference compared to the traditional LIF design.

\end{document}